\documentclass[A4paper,sagej,authoryear]{sagej}
\usepackage{moreverb,url}
\usepackage{csquotes}
\usepackage{xcolor}
\usepackage{natbib}
\usepackage{algorithm}
\usepackage{algpseudocode}
\usepackage{array}
\usepackage{ragged2e}
\newcolumntype{L}[1]{>{\RaggedRight\arraybackslash}p{#1}}
\usepackage[most]{tcolorbox}
\usepackage{makecell}
\makeatletter

\renewcommand{\@maketitle}{%
  \vspace*{0pt}
  \begin{center}
    {\titlesize\bfseries \@title\par}
    \vspace{1.5em}
    {\large\bfseries \@author\par}
  \end{center}
  \vspace{1.5em}
  \noindent\usebox\absbox\par
  \vspace{1em}
  \noindent\@keywords\par
  \vspace{1.5em}
}

\def\ps@title{%
  \let\@oddhead\@empty
  \let\@evenhead\@empty
  \def\@oddfoot{\hfil\thepage\hfil}
  \let\@evenfoot\@oddfoot
}

\makeatother
\definecolor{examplegray}{RGB}{244,244,244}
\definecolor{exampleborder}{RGB}{150,150,150}
\definecolor{exampledark}{RGB}{235,235,235}
\usepackage[colorlinks,bookmarksopen,bookmarksnumbered,citecolor=red,urlcolor=red]{hyperref}

\newcommand\BibTeX{{\rmfamily B\kern-.05em \textsc{i\kern-.025em b}\kern-.08em
T\kern-.1667em\lower.7ex\hbox{E}\kern-.125emX}}

\def\volumeyear{2026}

\begin{document}

\runninghead{The PTI procedure for mixed methods text analysis}

\title{From Qualitative Text to Integrated Evidence: The Parallel-Then-Integrated (PTI) Procedure}

%\author{Elyas Irankhah\affilnum{1} Ashley Hagaman\affilnum{2} and Samah Fodeh\affilnum{3}}

%\author{Anonymized\affilnum{1} and Anonymized\affilnum{2}}

%\affiliation{\affilnum{1}[Author affiliation removed for anonymized review]\\
%\affilnum{2}[Author affiliation removed for anonymized review]}

%\corrauth{Anonymized}

%\email{Anonymized}

\author{
Elyas Irankhah\affilnum{1},
Ashley Hagaman\affilnum{2}
and Samah Fodeh\affilnum{1}
}

\affiliation{
\affilnum{1}Department of Emergency Medicine,
Yale School of Medicine, Yale University,
New Haven, CT, USA\\
\affilnum{2}Department of Social and Behavioral Sciences,
Yale School of Public Health, Yale University,
New Haven, CT, USA
}

\corrauth{Samah Fodeh}

\email{samah.fodeh@yale.edu}

\begin{abstract}
Mixed methods research offers established strategies for integrating qualitative and quantitative evidence, but less guidance for integrating two computational representations of the same qualitative corpus. This methodological study introduces Parallel-Then-Integrated (PTI), a four-step procedure for explicit and traceable integration. We present PTI using 1,203 interview paragraphs from 29 mothers in Nepal. The two representations were ontology-guided Patient Voice (PV) coding with PVminer and data-driven themes generated through non-negative matrix factorization (NMF). A theme-complete crosswalk documented convergence, divergence, and preprocessing-sensitive differences. Depression status, based on Patient Health Questionnaire-9 (PHQ-9) scores, served as a secondary external anchor for exploratory association checks. PTI provides a structured approach for integrating computational representations of qualitative text within mixed methods research.
\end{abstract}

\maketitle

\section{Introduction}
Recent advances in large language models (LLMs) and computational text analysis have expanded how qualitative data can be analyzed within mixed methods research \citep{schroeder2025llm, guenduez2026patterns}. LLMs can support codebook-based deductive coding, identify data-driven patterns, and assist researchers in working systematically with larger bodies of qualitative text \citep{tai2024llm, mathis2024inductive, mannstadt2024novel}. These capabilities can reduce some of the manual burden of text analysis, support consistent application of structured coding frameworks across larger corpora, and help researchers examine patterns that would be difficult to review manually, while still requiring human interpretation and validation \citep{dunivin2025scaling, zeng2026advancing}. Mixed methods research already provides strategies for connecting qualitative and quantitative evidence. Qualitative data may be transformed into quantitative variables \citep{sandelowski2009quantitizing}, organized within joint displays \citep{guetterman2015joint}, or integrated to generate meta-inferences \citep{teddlie2009foundations}. As computational methods increasingly transform qualitative text into quantitative representations, researchers also need to make clear how those representations relate to participant meaning and to one another \citep{grimmer2013text, roberts2014structural, zhang2023redefining}. Two contrasting computational approaches are particularly relevant: data-driven theme discovery and ontology-guided coding. Unsupervised methods such as non-negative matrix factorization (NMF) identify patterns from participant language that do not automatically correspond to predefined clinical or social constructs \citep{lee1999learning, miner2023comparing}. Ontology-guided coding takes a different approach by mapping text to predefined conceptual categories \citep{garla2012ontology}. These approaches parallel inductive and deductive qualitative analysis: one develops patterns from the data, while the other begins with an existing conceptual framework \citep{proudfoot2023hybrid}. Their comparison can also involve abductive reasoning, because empirical patterns are considered in relation to existing concepts when developing interpretations \citep{timmermans2012theory}. The challenge is that two computational representations generated from the same qualitative corpus are not automatically comparable. Existing mixed methods frameworks establish principles for integration and document challenges in connecting findings across analytic components \citep{bryman2007barriers, ocathain2007integration}. Less guidance is available for determining how multiple computational representations of the same qualitative corpus should first be aligned, where they converge or diverge, and how those relationships should inform interpretation. We address this methodological gap with the \textit{Parallel-Then-Integrated} (PTI) procedure, a structured approach for preserving two computational representations separately before aligning and interpreting them together. PTI is not proposed as a theory-building framework, its purpose is to make the integration process explicit and traceable. It consists of four steps: parallel quantitizing, cross-stream alignment, assessment of overlap \& uniqueness, and integrated interpretation through a joint display.

%###################################################################
\subsection{Objectives}\label{subsec:aims}

The methodological aim of this study is to develop and illustrate PTI as a procedure for integrating multiple computational representations of the same qualitative corpus within mixed methods research. We apply PTI to participant-response text from a community-based perinatal mental health study \citep{vanheerden2024standstrong} and evaluate how two computational representations ontology-guided Patient Voice (PV) coding and data-driven non-negative matrix factorization (NMF) contribute to integrated interpretation. The analysis includes 1,203 participant-response paragraphs from 29 mothers. Patient Health Questionnaire-9 (PHQ-9) score-based depression status is used only as a mother-level external anchor for exploratory association checks. Accordingly, the demonstration addresses two questions: (1) How do ontology-guided Patient Voice coding and data-driven NMF contribute overlapping, distinct, and potentially unstable information when applied to the same qualitative corpus? (2) As a secondary association check, how do the individual representations and their combination relate to PHQ-9-based depression status?

\section{Mixed Methods Integration in Computational Text Analysis}

Contemporary mixed methods integration techniques provide useful starting points for connecting findings from different analytic components. Bringing multiple analyses into the same study does not, by itself, establish integration \citep{ocathain2007integration}. Researchers need to show how findings are connected and how each contributes to the overall study purpose \citep{bryman2006integrating, bazeley2009integrating, morgan2014integrating}. The triangulation protocol classifies relationships as agreement, partial agreement, silence, or disagreement. Following a thread traces findings across study components, while the mixed methods matrix organizes evidence within a shared case structure \citep{ocathain2010three}. Applying these approaches to computational text outputs introduces an additional problem: the outputs being compared may have different conceptual and numerical forms. For example, an ontology-guided method may assign a predefined code to a paragraph, while a data-driven model may assign a continuous numerical weight indicating how strongly that paragraph is associated with a particular theme \citep{baumer2017comparing, jacobs2019topic}. Placing the two values together identifies their common source, but does not establish that they represent the same concept. Joint displays can organize these relationships and support integrated conclusions \citep{fetters2022joint, haynesbrown2021jointdisplay}. Interpreting these relationships still depends on earlier decisions about which outputs should be aligned and what constitutes a meaningful correspondence \citep{peters2024missed}. A documented alignment stage can make these decisions transparent before integrated interpretation \citep{andreotta2019analyzing}. Related computational approaches within qualitative research also emphasize the need to connect model outputs back to source text. Computational grounded theory and multi-method text analysis combine pattern detection with close examination of qualitative material \citep{nelson2020computational, andreotta2019analyzing}. Quali-quantitative approaches similarly demonstrate the value of combining computational and interpretive analyses without treating their outputs as interchangeable \citep{blok2014complementary}. Computational methods can identify recurring patterns across larger collections of text, but researchers remain responsible for interpreting those patterns in relation to participant meaning and substantive questions \citep{guetterman2018augmenting}. Topic modeling illustrates this interpretive process. A topic is a weighted pattern of terms, where larger term weights indicate that particular words or phrases contribute more strongly to that topic \citep{jacobs2019topic}. Paragraphs can also receive topic weights indicating how strongly a topic is represented in the text. For example, a topic with highly weighted terms such as \textit{mother}, \textit{in-law}, \textit{sister}, and \textit{husband} may suggest family-related talk, but interpretation still requires examination of the participant paragraphs in which the topic is strongly represented \citep{towler2023applying, melnikov2024topic}. Analysts can also examine sensitivity to preprocessing choices when interpreting computational themes \citep{gillings2023interpretation}. Comparisons of topic modeling and human coding further show that the same qualitative material may be organized differently across analytic approaches \citep{miner2023comparing, gillies2022theme}. Machine-assisted quantitizing extends this issue by transforming qualitative material into numeric variables that can enter statistical or mixed methods analyses \citep{karjus2023machine}. Quantitizing makes it possible to summarize and compare patterns across a corpus, but computational representations may still differ in conceptual meaning, structure, and scale \citep{baumer2017comparing}. Current approaches such as the Pillar Integration Process (PIP) and the triangulation protocol provide structures for integrating or comparing findings \citep{johnson2019pillar, ocathain2010three}. These approaches were not developed specifically to operationalize the pre-integration alignment of multiple computational representations generated from the same qualitative corpus. PTI focuses on this narrower problem by making conceptual correspondence, unmatched outputs, and preprocessing-sensitive divergence explicit before joint interpretation.

%####################################################################

\section{Methods}
\subsection{Parallel-Then-Integrated (PTI) Procedure}

PTI organizes two computational representations of qualitative text and integrates them through four steps as follows:

\begin{enumerate}
\item[(1)] \textit{Parallel quantitizing:} Apply ontology-guided and data-driven methods independently to the same corpus, producing separate numeric representations \citep{karjus2023machine}.

\item[(2)] \textit{Cross-stream alignment:} Map the outputs of both representations to shared conceptual domains so that their relationships can be examined.

\item[(3)] \textit{Assessment of overlap and uniqueness:} Examine where the representations converge, diverge, or contribute distinct information.

\item[(4)] \textit{Integrated interpretation:} Synthesize the aligned evidence through a joint display to generate meta-inferences.
\end{enumerate}

To make the relationship between the two representations explicit and traceable, PTI uses four heuristic checks: domain uniqueness, feature overlap, signal increment, and interpretive richness (Table~\ref{tab:complementarity-criteria}). Each check captures a different aspect of the relationship, and the checks are considered separately without producing a single score or decision threshold. Here, domains refer to conceptual areas represented by a code or theme, features are the numeric variables derived from each representation, signal refers to association with the optional external anchor, and insights refer to interpretations that emerge when the two representations are considered together.

\begin{table}[h]
\small\sf\centering
\caption{Heuristic checks for examining relationships between PTI representations}
\label{tab:complementarity-criteria}
\begin{tabular}{p{0.22\textwidth}p{0.38\textwidth}p{0.26\textwidth}}
\toprule
Check & What it examines & Illustrative evidence \\
\midrule

Domain uniqueness
& Domains represented by only one representation
& Representation-specific domains \\

\addlinespace

Feature overlap
& Association between features from the two representations
& Cross-representation associations \\

\addlinespace

Signal increment
& Added signal when an external anchor is available
& Comparison with the individual representations \\

\addlinespace

Interpretive richness
& Insights produced through integrated interpretation
& Meta-inferences drawing on both representations \\

\bottomrule
\end{tabular}
\end{table}

Algorithm~\ref{alg:pti} presents the four PTI steps as a reusable analytic sequence. Specific coding and topic-modeling methods can vary across applications.

\begin{algorithm}[h]
\caption{Parallel-Then-Integrated (PTI) procedure}
\label{alg:pti}
\begin{algorithmic}

\Require Text corpus $X$; ontology-guided method $O$; data-driven text-analysis method $T$; optional external anchor $Y$
\Ensure Cross-stream crosswalk $C$; joint display $J$; meta-inferences $M$

\Statex
\Statex \textbf{Step 1: Parallel quantitizing}
\State Apply $O$ to each text unit and retain codes and supporting text spans
\State Apply $T$ independently to the same text units and retain theme weights
\State Aggregate both representations to the common analytic unit

\Statex
\Statex \textbf{Step 2: Cross-stream alignment}
\State Map ontology codes and themes to shared conceptual domains
\State Construct a theme-complete crosswalk $C$, retaining unmatched domains

\Statex
\Statex \textbf{Step 3: Assessment of overlap and uniqueness}
\State Assess domain uniqueness and feature overlap
\If{$Y$ is available}
    \State Assess signal increment against $Y$
\EndIf
\State Record supporting evidence for each assessment

\Statex
\Statex \textbf{Step 4: Integrated interpretation}
\State Construct joint display $J$ from the aligned evidence
\State Identify convergence, divergence, and representation-specific information
\State Generate meta-inferences $M$ from the integrated evidence
\State Assess interpretive richness

\Statex
\State Document applicable boundary conditions
\State \Return $C, J, M$

\end{algorithmic}
\end{algorithm}

%####################################################################

\subsection{Data Source}\label{subsec:design-data}

The analysis uses de-identified interview transcripts from a community-based perinatal mental health study of adolescent and young mothers in Chitwan, Nepal \citep{vanheerden2024standstrong}. Interviews were conducted in Nepali, transcribed verbatim in Nepali, and translated into English, preserving culturally meaningful terms \citep{poudyal2019wearable}. The interviews elicited experiences of young motherhood and passive-sensing technology. Participants were recruited through postnatal clinics and immunization camps. The analytic corpus included only participant responses, interviewer questions, template text, and non-speech metadata were excluded. 

%\subsection{Analytic cohort, outcome, and text units}\label{subsec:cohort-outcome-text}
\subsection{Study Design}\label{subsec:cohort-outcome-text}

The illustrative analytic sample comprised $n = 29$ mothers from the parent Nepal study with participant-response text and binary depression status (9 depressed, 20 non-depressed). Here, the external anchor refers to depression status measured independently of the two text representations. Depression status was defined from the PHQ-9 screening score using a cutoff of $\geq 9$, following the parent study's Chitwan-based validation approach \citep{vanheerden2024standstrong}. Paragraphs were coded at the text level and aggregated to mother-level features before modeling. The binary outcome was defined at the mother level. Let $y_i \in \{0,1\}$ denote depression status for mother $i$, where $y_i = 1$ indicates depressed and $y_i = 0$ indicates non-depressed:

\begin{equation}
y_i =
\begin{cases}
1, & \text{if mother } i \text{ is depressed}, \\
0, & \text{if mother } i \text{ is non-depressed}.
\end{cases}
\end{equation}

The 4,253 transcript rows included participant responses, interviewer questions, and other non-analytic text. Because PVminer and NMF were applied only to participant responses, interviewer and other non-participant text were excluded. This left 1,242 participant-response paragraphs. We then applied minimal normalization by stripping leading speaker tags and bullet markers and collapsing whitespace. We also removed very short turns (fewer than five alphabetic words after normalization) to exclude acknowledgments and backchannels that carry little semantic content for coding or topic modeling. This second step removed 39 rows (3.1\%), yielding 1,203 analytic paragraphs. Table~\ref{tab:text-units} presents the preprocessing funnel. Per-mother paragraph counts reflect retained participant text only. Let $m_i$ denote the number of retained paragraphs for mother $i$, and let $p_{ij}$ denote paragraph $j$ from mother $i$, where $j = 1,\ldots,m_i$. All PV labels and NMF theme weights were aggregated to mother-level vectors before modeling.

\begin{figure}[htbp]
\centering
\includegraphics[width=0.95\textwidth]{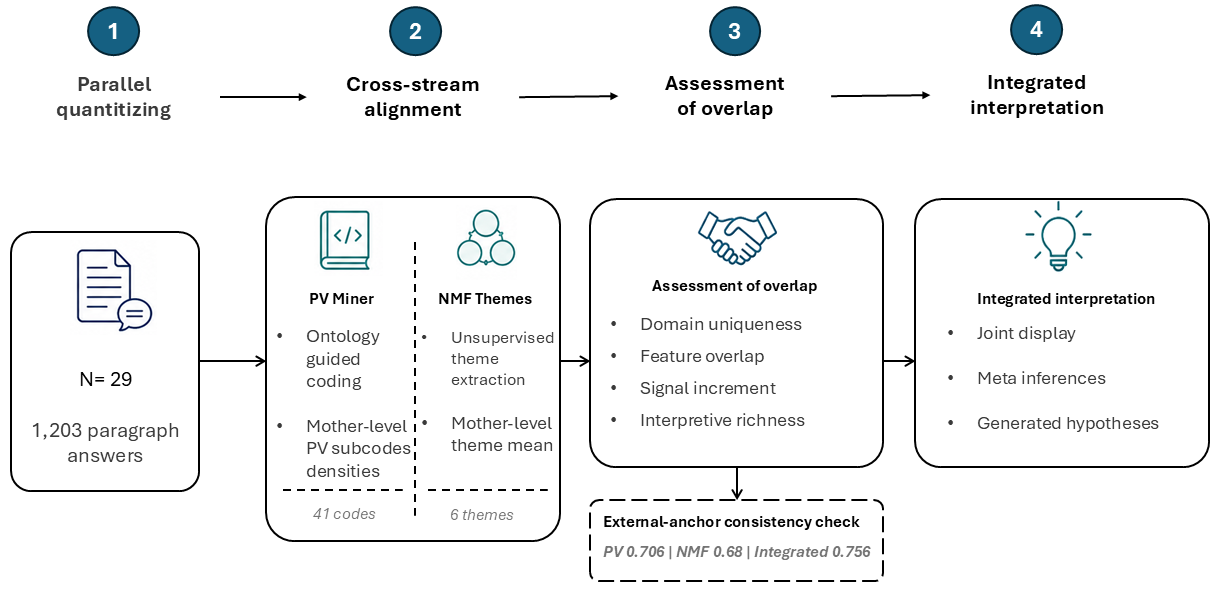}
\caption{Overview of the PTI demonstration. Participant-response paragraphs are transformed into two independent computational representations (PVminer and NMF), then aligned, assessed for overlap, and integrated through the joint display. Cross-validated logistic regression provides a secondary external-anchor association check. The joint display is the primary analytic output.}
\label{fig:workflow}
\end{figure}

\begin{table}[htbp]
\small\sf\centering
\caption{Preprocessing funnel: transcript rows to analytic paragraphs}\label{tab:text-units}
\begin{tabular}{lrrrrrr}
\toprule
Stage & Mothers & Rows & Removed & \multicolumn{3}{c}{Per-mother counts} \\
\cmidrule(lr){5-7}
 & & & & Min & Mean & Max \\
\midrule
29-mother analytic cohort & 29 & 4,253 & --- & 83 & 146.7 & 233 \\
Participant turns only & 29 & 1,242 & 3,011 & 12 & 42.8 & 95 \\
After short-turn filter ($\geq$5 words) & 29 & 1,203 & 39 & 12 & 41.5 & 93 \\
\bottomrule
\end{tabular}
\end{table}
\noindent\textit{Note:} Per-mother counts are reported at each stage. The 3,011 rows removed at the participant-turn stage comprise interviewer turns (2,076), boilerplate/field-note text (55), and other non-participant or unlabeled rows (880). The short-turn filter removed 39 paragraphs (3.1\%); all 29 mothers retained at least 12 paragraphs. Figure~\ref{fig:workflow} summarizes the demonstration workflow. It shows how participant responses are transformed into parallel PVminer and NMF representations, then aligned and integrated through PTI.

\subsection{PVminer codebook and feature construction}\label{subsec:pv-tabpo}

The ontology-guided representation used PVminer to assign Patient Voice ontology labels to participant-response paragraphs \citep{fodeh2026tabpo}. The Patient Voice ontology organizes patient and provider language into predefined communication and psychosocial domains. One of its top-level codes is Social Determinants of Health (SDOH), which includes sub-codes representing areas such as economic stability, neighborhood and built environment, and social and community context. These ontology-defined codes form the PV representation that is later compared with the data-driven NMF themes to address the first research question. We used the Token-Adaptive Barrier Preference Optimization (TAB-PO) Llama~3.3 70B checkpoint for all reported PV labels \citep{fodeh2026tabpo}. The 70B checkpoint had the strongest reported ontology-consistent labeling performance on the PVminer expert gold standard (approximately 85\% code $F_1$, 83\% sub-code $F_1$, and 88\% span $F_1$). Our PV features depend on stable, schema-valid \texttt{Code}, \texttt{Sub-code} assignments across diverse participant-response paragraphs. Here, schema-valid means that each predicted \texttt{Sub-code} is permitted under its assigned \texttt{Code} according to the Patient Voice ontology. In development runs on this corpus, smaller checkpoints produced lower schema-valid yield and overbroad sub-code assignments (e.g., SDOH and care coordination codes appearing in nearly all participants), which would weaken the usefulness of the mother-level density features, defined as the proportion of each mother's paragraphs assigned to a given \texttt{Code}, \texttt{Sub-code} pair. TAB-PO maps free text to schema-valid PV records. Each record contains a coarse-grained \texttt{Code}, a fine-grained \texttt{Sub-code} valid under that code, and a grounded text span from the source paragraph. For a paragraph $p_{ij}$ from mother $i$, PVminer returns a set of PV records:

\begin{equation}
L_{ij} = \{(\kappa,\varsigma,\pi)\},
\end{equation}
where $\kappa$ denotes the PV code, $\varsigma$ denotes the corresponding sub-code, and $\pi$ denotes the supporting text span. The ontology was developed for structured labeling of patient and provider utterances in clinical dialogue. Table~\ref{tab:pv-codebook} summarizes the PV domains and sub-codes represented in the present analysis.

\begin{table}[h]
\small\sf\centering
\caption{Patient Voice ontology domains}
\label{tab:pv-codebook}
\begin{tabular}{p{0.32\textwidth}p{0.58\textwidth}}
\toprule
PV domain & Content focus/sub-codes \\
\midrule

Care coordination
& Patient/provider coordination of care \\

\midrule

Partnership
& Active participation/involvement, express opinions, state preferences, alignment, appreciation/gratitude, connection, salutation, build trust \\

\midrule

Shared decision-making
& Explore options, seeking approval, approval of decision, share options, make decision, approval/reinforcement \\

\midrule

SDOH
& Economic stability, education access and quality, health care access and quality, neighborhood and built environment, social and community context \\

\midrule

Socioemotional behavior
& Express emotions/concern/unease, sadness/fear, acknowledge situation, reassurance, positive remarks/reward \\

\bottomrule
\end{tabular}
\end{table}

The ontology has no dedicated infant-care or baby care construct. Infant-care talk was assigned to existing PV categories such as care coordination or \texttt{SDOH}/\texttt{SocialAndCommunityContext} when TAB-PO returned schema-valid PV labels. For the present external-anchor analysis, we used the semantic \texttt{Code}--\texttt{Sub-code} pair as the feature source. Grounded spans linked each label to the paragraph text but were not entered as predictors. TAB-PO assigned labels across the 1,203 analytic paragraphs. Forty-one distinct \texttt{Code}--\texttt{Sub-code} pairs appeared at least once in the analytic sample ($|\mathcal{S}| = 41$). Paragraph-level PV labels were aggregated to mother-level density features so that the unit of analysis matched the mother-level depression outcome. Let $\mathcal{S}$ denote the set of observed PV \texttt{Code}--\texttt{Sub-code} pairs. For mother $i$ and PV pair $s \in \mathcal{S}$, the density feature was defined as:

\begin{equation}
x_{is}^{\mathrm{PV}}
=
\frac{1}{m_i}
\sum_{j=1}^{m_i}
\mathbb{I}\left(s \in L_{ij}^{*}\right),
\end{equation}

where $m_i$ is the number of participant-response paragraphs for mother $i$, $L_{ij}^{*}$ is the set of \texttt{Code}, \texttt{Sub-code} pairs assigned to paragraph $p_{ij}$, and $\mathbb{I}(\cdot)$ is the indicator function. Therefore, $x_{is}^{\mathrm{PV}}$ represents the proportion of mother $i$'s response paragraphs assigned PV pair $s$. This density representation serves two purposes. First, it normalizes for unequal interview length across mothers. Second, it creates a mother-level feature matrix for leakage-aware cross-validation. The result is a $29 \times 41$ candidate matrix of sub-code-density features.

%##############################################################
\subsection{PV external-anchor association model}\label{subsec:primary-pv-model}

The PV model tested whether ontology-guided sub-code-density features were associated with mother-level depression status under leave-one-mother-out cross-validation (LOOCV). The five PV sub-codes used in each model were not fixed in advance. Instead, feature selection was repeated separately within each LOOCV training fold, using only the training mothers. The held-out mother's outcome was not used to select features or fit the model. For each fold, one mother was held out. Using only the remaining 28 mothers in the training set, each of the 41 sub-code-density features was ranked by its univariate association with depression status in the training set (the absolute distance between the training-fold area under the receiver operating characteristic curve and 0.5). The top five sub-codes were retained for that fold, and $\ell_2$-regularized logistic regression was fit on the resulting $28 \times 5$ training matrix:
\begin{equation}
\Pr(y_i=1 \mid \mathbf{x}^{\mathrm{PV}}_i)
=
\sigma\left(\beta_0 + {\mathbf{x}^{\mathrm{PV}}_i}^{\top}\boldsymbol{\beta}\right),
\end{equation}
where $\mathbf{x}^{\mathrm{PV}}_i$ denotes the vector of five fold-selected PV sub-code densities for mother $i$, and $\sigma(z) = 1/(1+\exp(-z))$ is the logistic function. The fitted model generated one predicted probability for the held-out mother. Because feature selection was repeated inside each training fold, the held-out mother's outcome did not influence which sub-codes entered the model. For the external-anchor analysis, association was summarized using the area under the receiver operating characteristic curve (AUC). The analysis was designed to assess external-anchor association, not diagnostic performance. We use the term external-anchor association because discrimination against a single binary criterion (depression status) does not, on its own, establish the convergent, discriminant, and nomological evidence that construct validation requires. PVminer does not diagnose depression. It quantifies codebook-based psychosocial domains for mixed methods research.

\subsection{Unsupervised NMF theme representation}

NMF was used to quantify qualitative themes from participant-response paragraphs without imposing the PV ontology. A theme is the main topic the participant mentioned in her response. This data-driven representation provides the essential comparison: PVminer ontology-guided features versus NMF data-driven theme features derived from the same participant responses. NMF does not replace qualitative interpretation. It offers a parallel numeric representation that can be aligned with other computational representations within mixed methods research. The NMF comparator applied unsupervised topic modeling to the same 1,203 participant-response paragraphs used for PVminer labeling. We use \emph{theme} for an interpreted NMF component and \emph{topic modeling} for the computational method. Features included unigrams and bigrams with English stop-word removal and document-frequency filtering. As a preprocessing sensitivity check, we repeated the $k = 6$ NMF after contraction normalization, expanding common English contractions (e.g., \enquote{don't} $\rightarrow$ \enquote{do not}) while keeping the remaining preprocessing and model settings unchanged. This check examined whether theme interpretation was sensitive to contraction fragments. Let $\mathbf{V} \in \mathbb{R}_{+}^{D \times T}$ denote the term frequency--inverse document frequency (TF--IDF) matrix for $D = 1{,}203$ participant-response paragraphs and $T$ vocabulary terms. NMF approximated this matrix as:
\begin{equation}
\mathbf{V} \approx \mathbf{W}\mathbf{H}
\end{equation}
where $\mathbf{W} \in \mathbb{R}_{+}^{D \times k}$ contains paragraph-level theme weights and $\mathbf{H} \in \mathbb{R}_{+}^{k \times T}$ contains theme-term weights. Bootstrap solution stability selected $k = 6$ from $k = 2$ to $k = 8$ by maximizing the mean Jaccard similarity of top-term matches across bootstrap refits (mean Jaccard = 0.410). NMF was fit on the full 1,203-paragraph corpus before cross-validation. The resulting theme weights are transductive. The held-out mother's paragraphs contributed to the unsupervised representation, although her depression status was not used. Only supervised feature selection and logistic-regression fitting were fold specific. Standalone NMF and combined PV--NMF AUCs should therefore be interpreted as exploratory transductive estimates, not as fully out-of-sample performance. Paragraph-level theme weights were aggregated to the mother level using mean theme weights for the standalone NMF comparator. For mother $i$ and theme $t$, the feature was

\begin{equation}
\bar{w}_{it}
=
\frac{1}{m_i}
\sum_{j=1}^{m_i}
w_{ijt},
\quad t=1,\ldots,6,
\end{equation}
where $w_{ijt}$ is the NMF weight for theme $t$ in paragraph $j$ from mother $i$. This produced a $29 \times 6$ mother-level comparator matrix for the standalone NMF endpoint. The combined model used the same six participant-response NMF themes, summarized as mother-level theme features for fold-specific block selection. Table~\ref{tab:ashley-themes} summarizes the six themes.

\begin{table}[h]
\small\sf\centering
\caption{Participant-response NMF themes ($k = 6$)}
\label{tab:ashley-themes}
\begin{tabular}{p{0.07\textwidth}p{0.20\textwidth}p{0.30\textwidth}p{0.35\textwidth}}
\toprule
Theme & Label & Content focus & Top TF--IDF terms \\
\midrule

T0 & Baby care and health 
& Infant care, monitoring illness, and providing care 
& baby, care, care baby, look, look baby, sick, birth, provide \\
\midrule

T1 & Emotional difficulty and help-seeking 
& Feeling difficulty and asking for or receiving help 
& feel, difficult, feel difficult, didn, didn feel, help, don feel, ask \\
\midrule

T2 & Activities and preferences 
& Leisure, play, likes, and dislikes 
& like, things, play, don like, talk, activities like, works, activities \\
\midrule

T3 & Family and in-laws 
& Husband, parents, sisters, and extended family 
& mother, in-law, mother in-law, sister, husband, sister in-law, father, father mother \\
\midrule

T4 & Denial and negation 
& Denying problems, uncertainty, and negation talk 
& don, problems, don problems, umm, think, know, don know, things \\
\midrule

T5 & Daily chores and hygiene 
& Washing, clothing, food, and household time use 
& time, clothes, used, wash, wash clothes, washing, food, washing clothes \\
\bottomrule
\end{tabular}
\end{table}

\subsection{Cross-stream alignment and joint-display decisions}

Alignment used PV ontology definitions, NMF terms, representative paragraphs, and grounded PV spans. Convergence was used when both representations reflected the same general domain. Partial convergence was used when they overlapped but emphasized different aspects of the text. Divergence was used when no clear match was found or when the relationship did not hold after the preprocessing sensitivity check. All themes were kept in the crosswalk. A theme was included in the joint display when it added a clear and useful perspective. Themes with weak, indirect, or preprocessing-sensitive links were kept in the crosswalk but not included in the joint display. For each theme, the five paragraphs with the highest theme weights were reviewed.

%\subsection{AI-assisted robustness audit}\label{subsec:ai-audit}

%We conducted a blinded multi-model audit of the PV assignments and cross-stream alignment. An initial review examined 50 PV-coded paragraphs, with 20 independently reviewed by a second model as an overlap set for agreement assessment. The overlap set comprised the first 20 sampled paragraphs ordered by record identifier. OpenAI GPT-5.6 Sol and Anthropic Claude Opus 5 reviewed the overlap set and all six NMF-to-PV mappings, while Google Gemini 3.8 Flash adjudicated disagreements. PV assignments were rated as consistent, partial, or inconsistent, and theme mappings were rated for alignment and joint-display inclusion. The audit served as a secondary robustness check.

\subsection{Combined PV--NMF external-anchor model}\label{subsec:integrated-model}

The combined model tested whether combining ontology-guided and data-driven participant-response transformations provided incremental external-anchor signal beyond either representation alone. It concatenated fold-selected PV features and fold-selected NMF theme features in a single mother-level logistic regression model:

\begin{equation}
\mathbf{x}_i =
\left[ \mathbf{x}_i^{\mathrm{PV}} \,\|\, \mathbf{x}_i^{\mathrm{NMF}} \right]
\end{equation}
\noindent where $\|$ denotes feature concatenation, and $\mathbf{x}_i^{\mathrm{PV}}$ and $\mathbf{x}_i^{\mathrm{NMF}}$ represent the selected PV and NMF features for mother $i$, respectively. The resulting vector was used as input to the logistic regression model described in the Statistical Modeling subsection. Within each LOOCV training fold, the top five PV sub-codes were selected from the pool of 41 sub-codes, and up to four participant-response NMF theme features were selected from the six themes in Table~\ref{tab:ashley-themes}. These NMF features came from the same $k = 6$ participant-response NMF solution used for the standalone comparator. Non-informative features with insufficient training-fold variation were removed before ranking, so the number of NMF features entering the combined model could vary across folds. The effective NMF contribution to the combined model was two theme features (T0, baby care and health; and T2, activities and preferences), even when the NMF block cap was expanded from four to six themes. Selected PV and NMF features were combined in a single matrix. The combined model provides a secondary external-anchor representation by entering the concatenated PV and NMF features in one logistic regression.

\subsection{Statistical modeling}\label{subsec:modeling}

 In this analysis we compare the three data representations: the combined PV-NMF external anchor, PVminer features, and NMF themes. All evaluations were performed at the mother level. Paragraphs were never treated as independent participants. For each paragraph representation, text-derived features were first aggregated to one row per mother and then entered into a regularized logistic regression model. The three models used this same mother-level evaluation so that they are directly comparable. Supervised model predictions were evaluated using held-out mothers under cross-validation. Let $\mathbf{x}_i \in \mathbb{R}^{q}$ denote the mother-level feature vector for mother $i$, where $q$ depends on the representation being evaluated. The binary depression outcome was modeled as
\begin{equation}
\Pr(y_i=1 \mid \mathbf{x}_i)
=
\sigma\left(\beta_0 + \mathbf{x}_i^{\top}\boldsymbol{\beta}\right),
\end{equation}

where $\sigma(z) = 1/(1+\exp(-z))$ is the logistic function. Unless otherwise stated, models used $\ell_2$-regularized logistic regression with the default regularization setting. Regularization was used because the cohort was small and the number of depressed cases was limited. The same modeling approach was applied to PVminer features and NMF themes. Models were not interpreted as clinical screening tools; they assessed preliminary external-anchor association at the mother level. Primary evaluation used leave-one-mother-out cross-validation (LOOCV). In each fold, one mother was held out as the test case, all model fitting was performed using the remaining $n - 1$ mothers, and a predicted probability was generated for the held-out mother. The final performance estimate was computed from the set of held-out predictions across all 29 folds. This design limited supervised information leakage and avoided inflating the effective sample size. The main performance metric was AUC, computed from cross-validated mother-level predicted probabilities. Let $\hat{p}_{i}$ denote the held-out predicted probability for mother $i$. The AUC can be interpreted as
\begin{equation}
\operatorname{AUC}
=
\Pr\left(\hat{p}_{i^+} > \hat{p}_{i^-}\right),
\end{equation}
where $i^+$ and $i^-$ denote randomly selected depressed and non-depressed mothers, respectively. Permutation testing was used to assess whether the observed AUC exceeded chance. We used $B=500$ permutations. For each permutation $b = 1,\ldots,B$, mother-level depression labels were randomly shuffled, the full cross-validation procedure was rerun, and a permuted AUC was computed. The one-sided permutation $p$-value was calculated as:
\begin{equation}
p_{\mathrm{perm}}
=
\frac{
1 + \sum_{b=1}^{B}
\mathbb{I}\left(
\operatorname{AUC}^{(b)}_{\mathrm{perm}}
\geq
\operatorname{AUC}_{\mathrm{obs}}
\right)
}{
1 + B
}.
\end{equation}

%###########################################################################

\section{Results}\label{sec:results}

Results are organized around the analytic logic of PTI. We first examine what each representation captured independently, then how the representations aligned or diverged, how overlap and uniqueness were assessed, and what meta-inferences and hypotheses emerged from integration. Conceptual and interpretive evidence form the primary basis of the assessment, while cross-validated performance serves only as a secondary external-anchor association check. Table~\ref{tab:main-results} shows the illustrative performance results used in this assessment. The standalone PV model used five sub-codes selected within each LOOCV training fold from the pool of 41 sub-codes. The combined model capped selection at five PV sub-codes and four NMF themes per fold; its effective contribution was five PV and two NMF features. NMF themes were discovered on the full 1,203-paragraph corpus before LOOCV, so the NMF and combined AUCs are transductive estimates.

\begin{table}[h]
\small\sf\centering
\caption{LOOCV AUC by representation ($n = 29$ mothers)}\label{tab:main-results}
\begin{tabular}{p{0.32\textwidth}p{0.18\textwidth}p{0.12\textwidth}cc}
\toprule
Representation & Features & Evaluation & AUC & Permutation $p$ \\
\midrule
PV model (standalone) & 5 & LOOCV & 0.706 & 0.106 \\
NMF themes (standalone) & 6 & LOOCV & 0.680 & 0.090 \\
Combined PV + NMF & 5 PV + up to 4 NMF & LOOCV & 0.756 & 0.050 \\
\bottomrule
\end{tabular}
\end{table}

\subsection{What ontology-guided PV coding captured}\label{subsec:res-pv}

PV coding captured psychosocial domains defined in the ontology (Table~\ref{tab:pv-codebook}). Partnership codes captured relational and communication experiences, while SDOH codes captured economic, environmental, social, and community contexts. Each PV label was grounded in a supporting text span. As an external-anchor check, PV densities reached a LOOCV AUC of 0.706 (permutation $p = 0.106$). This indicated a directional association with depression status. The main contribution of the PV representation was its construct-level domain coverage.

\subsection{What data-driven NMF captured}\label{subsec:res-nmf}

NMF identified data-driven participant-language patterns, including activities and preferences (T2), denial and negation (T4), family and in-law talk (T3), and emotional difficulty and help-seeking (T1) (Table~\ref{tab:ashley-themes}). The six NMF themes reached a LOOCV AUC of 0.680 (permutation $p = 0.090$). T4 was sensitive to preprocessing. After contraction normalization, the denial and negation pattern did not replicate consistently. We therefore excluded T4 from integrated interpretation. T2 showed partial convergence with PV active participation and involvement, but its correspondence was indirect. Routine-disruption language was captured more clearly by T5. Accordingly, T5 was retained in the joint display, while T2 was retained only in the crosswalk.

%#########################################################

\subsection{Cross-stream alignment, assessment, and integrated interpretation}
\label{subsec:worked-example}
\label{subsec:res-integrated}

\newtcolorbox{workedexamplebox}[1][]{
    breakable,
    colback=exampledark,
    colframe=exampleborder,
    boxrule=0.7pt,
    arc=1.5mm,
    left=3mm,
    right=3mm,
    top=2mm,
    bottom=2mm,
    before skip=6pt,
    after skip=8pt,
    fonttitle=\bfseries,
    title={#1}
}

\newtcolorbox{exemplarbox}[1][]{
    breakable,
    colback=examplegray,
    colframe=exampleborder,
    boxrule=0.5pt,
    arc=1.5mm,
    left=3mm,
    right=3mm,
    top=2mm,
    bottom=2mm,
    before skip=5pt,
    after skip=5pt,
    fonttitle=\bfseries,
    title={#1}
}

The analysis proceeded from cross-stream alignment to an assessment of overlap and uniqueness, followed by integrated interpretation through the joint display.
%The AI-assisted audit showed exact agreement on 18 of 20 PV paragraph ratings ($\kappa=0.855$), adjudicated ratings were five consistent, seven partial, seven inconsistent, and one with no PV label. For the six theme-level mappings, the two models agreed on four alignment categories ($\kappa=0.538$) and four joint-display decisions ($\kappa=0.250$). After adjudication, the audit agreed with the analyst's joint-display inclusion decision for five of six themes. T0 was the only inclusion-level discrepancy.

\textit{Cross-stream alignment.}
Alignment drew on PV ontology definitions, NMF terms, representative paragraphs, and grounded PV spans. The following scaffold makes this reasoning explicit.

\begin{workedexamplebox}[PTI example scaffold]

\textbf{Participant account}

The source paragraph provides the qualitative material.

\vspace{4pt}

\textbf{PV representation}

\begin{quote}
\ttfamily\small
\{
\par\hspace{1em}"Code": "[ontology code]",
\par\hspace{1em}"Sub-code": "[ontology sub-code]",
\par\hspace{1em}"Span": "[grounded participant wording]"
\par\}
\end{quote}

\vspace{3pt}

\textbf{NMF interpretation}

The paragraph is related to one or more data-driven themes.

\vspace{3pt}

\textbf{Integrated interpretation}

The PV construct is related to the way the experience is expressed in the NMF theme.

\vspace{3pt}

\textbf{\textit{Meta-inference}}

The relationship between the two representations supports an integrated conclusion.

\end{workedexamplebox}

Two examples illustrate partial convergence between ontology-guided and data-driven representations in economic--relational and affective contexts.

\begin{exemplarbox}[Example 1: Economic--relational strain and violence (depressed)]

\textbf{\textit{Participant:}}
\enquote{My husband doesn't have any work to do, he doesn't take care of the house and this baby\ldots\ If I suggest him about such things, he starts to beat me.}

\vspace{3pt}

\textbf{\textit{PV representation:}}

\begin{quote}
\ttfamily\small
\{
\par\hspace{1em}"Code": "SDOH",
\par\hspace{1em}"Sub-code": "EconomicStability",
\par\hspace{1em}"Span": "My husband doesn't have any work to do"
\par\}
\end{quote}

\textbf{\textit{NMF interpretation:}}
The participant paragraph aligns with family and husband talk (T3).

\vspace{3pt}

\textbf{\textit{Integrated interpretation:}}
PV identifies economic instability, while NMF locates the experience within family and husband language. Their alignment shows that material hardship is narrated through a relational context that also includes violence.

\vspace{3pt}

\textbf{\textit{Meta-inference:}}
Economic hardship and intimate partner violence form a connected relational experience in this account. PTI documents how this relationship is divided across the two computational representations.

\end{exemplarbox}

\begin{exemplarbox}[Example 2: Socioemotional affect and emotional difficulty (non-depressed)]

\textbf{\textit{Participant:}}
\enquote{I feel irritate when he starts crying. I don't feel irritate when he has some problems but sometimes babies used to cry without any problems at that time I feel irritate.}

\vspace{3pt}

\textbf{\textit{PV representation:}}

\begin{quote}
\ttfamily\small
\{
\par\hspace{1em}"Code": "SocioEmotionalBehaviour",
\par\hspace{1em}"Sub-code": "expressEmotions",
\par\hspace{1em}"Span": "I feel irritate"
\par\}
\end{quote}

\textbf{\textit{NMF interpretation:}}
The participant paragraph aligns with emotional-difficulty and help-seeking talk (T1).

\vspace{3pt}

\textbf{\textit{Integrated interpretation:}}
PV assigns the affective content to the ontology's socioemotional construct without specifying emotional valence, while NMF places the same paragraph in the data-driven emotional-difficulty theme. The two representations therefore converge on emotional experience, while NMF additionally situates that experience within a difficulty-oriented context.

\vspace{3pt}

\textbf{\textit{Meta-inference:}}
Affective language about irritation and infant crying is captured by both representations, but at different levels. PV identifies socioemotional expression, while NMF characterizes the broader pattern as emotional difficulty. This represents partial cross-stream convergence, with NMF adding information about the negative or difficulty-oriented context of the emotional expression.

\end{exemplarbox}

Table~\ref{tab:crosswalk} maps all six NMF themes to relevant PV domains and records unmatched, excluded, and preprocessing-sensitive themes.

\begin{table*}[htbp]
\scriptsize\sf
\centering
\setlength{\tabcolsep}{4pt}
\renewcommand{\arraystretch}{1.22}

\caption{Theme-complete cross-stream crosswalk: NMF themes mapped to relevant PV domains, with joint-display inclusion decisions}
\label{tab:crosswalk}

\begin{tabular}{%
L{0.14\textwidth}
L{0.25\textwidth}
L{0.20\textwidth}
L{0.31\textwidth}
}
\toprule
\textbf{NMF theme} &
\textbf{Mapped PV domain(s)} &
\textbf{Alignment} &
\textbf{Joint-display decision} \\
\midrule

\textbf{T0:} Baby care and health
&
Care coordination; \texttt{SDOH}/\texttt{SocialAnd
CommunityContext}
&
Partial convergence; no dedicated infant-care construct
&
Excluded: already covered by existing care and social-context codes.
\\
\addlinespace
\midrule

\textbf{T1:} Emotional difficulty and help-seeking
&
Socioemotional behavior
&
Partial convergence
&
Included as \textit{Emotional difficulty}; NMF adds a difficulty-oriented context not specified by the PV construct.
\\
\addlinespace
\midrule

\textbf{T2:} Activities and preferences
&
Partnership/active participation and involvement
&
Partial convergence
&
Excluded from joint display: shared activity/engagement focus, but PV is care/interaction-oriented while T2 is everyday likes, play, and preferences; routine disruption maps to T5.
\\
\addlinespace
\midrule

\textbf{T3:} Family and in-laws
&
Partnership, \texttt{SDOH}/\texttt{EconomicStability}
&
Convergence with Partnership, partial convergence with \texttt{SDOH} &
Included as two joint-display rows: \textit{Economic hardship} and \textit{Partnership and family}.
\\
\addlinespace
\midrule

\textbf{T4:} Denial and negation
&
None
&
Apparent divergence under default preprocessing
&
Excluded: failed contraction-normalization sensitivity; retained as a methodological finding.
\\
\addlinespace
\midrule

\textbf{T5:} Daily chores and hygiene
&
None
&
Divergence: NMF-only
&
Included as \textit{Routine and daily-life disruption}.
\\

\bottomrule
\end{tabular}

\end{table*}

\textit{Assessment of overlap and uniqueness.}
Across Steps~3 and~4, PTI applies the four heuristic checks in Table~\ref{tab:complementarity-criteria}. Given the small sample, conceptual and interpretive evidence is primary, but the empirical checks remain exploratory. From the PV side, Shared decision-making had no direct NMF counterpart in the present crosswalk. T5 represented routine and daily-life disruption without a matching PV construct. T4 was excluded after failing the preprocessing sensitivity check. T2 showed only partial correspondence with PV active participation and was retained in the crosswalk but not the joint display. The mean absolute correlation between PV sub-code densities and NMF theme weights was $|r| = 0.152$ across 246 feature pairs and is reported descriptively. The combined PV--NMF model reached a LOOCV AUC of 0.756, compared with 0.706 for PV and 0.680 for NMF. This higher observed AUC is treated as exploratory evidence of partially non-overlapping external-anchor signal. Notably, the NMF features contributing to the combined external-anchor model were T0 and T2, both of which were excluded from the joint display. This shows that external-anchor signal and interpretive relevance were carried by different themes in this demonstration. The crosswalk and joint display provide the primary interpretive integration. Table~\ref{tab:joint-display} summarizes the resulting alignment and meta-inferences.

\begin{table*}[htbp]
\scriptsize\sf
\centering
\setlength{\tabcolsep}{3pt}
\renewcommand{\arraystretch}{1.08}

\caption{Joint display of cross-stream alignment and meta-inferences (PTI Steps~2--4)}
\label{tab:joint-display}

\begin{tabular}{%
p{0.12\textwidth}%
p{0.20\textwidth}%
p{0.05\textwidth}%
p{0.26\textwidth}%
p{0.27\textwidth}%
}
\toprule
\multicolumn{1}{c}{Domain} &
\multicolumn{1}{c}{PV code--sub-code} &
\multicolumn{1}{c}{NMF} &
\multicolumn{1}{c}{Alignment} &
\multicolumn{1}{c}{Meta-inference} \\
\midrule

Economic hardship
&
\shortstack[l]{%
\texttt{SDOH}/\\
\texttt{EconomicStability}}
&
T3
&
Partial convergence: NMF captures hardship through family and husband language.
&
Economic hardship is voiced relationally, not as isolated financial talk.
\\
\midrule

Partnership and family
&
\shortstack[l]{%
\texttt{PartnershipPatient}/\\
\texttt{expressOpinions}}
&
T3
&
Convergence: both representations capture relational content.
&
In the illustrated account, distress is voiced through conflict with the husband.
\\
\midrule

Emotional difficulty
&
\texttt{SocioEmotional
Behaviour}
&
T1
&
Partial convergence: both representations capture affective content, while NMF adds a difficulty-oriented context.
&
Distress is expressed through felt difficulty and help-seeking.
\\
\midrule

Routine and
daily-life disruption
&
None
&
T5
&
Divergence: NMF captures routine-disruption language absent from the ontology.
&
Routine-related language may be incompletely represented by the ontology, indicating a candidate coverage gap requiring validation.
\\

\bottomrule
\end{tabular}
\end{table*}
%#########################################################

\section{Discussion}\label{sec:discussion}

PTI was developed and illustrated as a procedure for integrating two computational representations of the same qualitative corpus. Its contribution is procedural: PTI makes explicit how independently generated outputs are aligned, compared, and interpreted together. This extends mixed methods attention to meaningful connections across analytic components \citep{small2011mixed} and complements recent work combining qualitative coding with computational topic modeling while preserving their distinct analytic contributions \citep{hoffman2025bigvape}. Applying PTI to the Nepal interview data showed that ontology-guided and data-driven approaches can represent the same participant-response text differently. Some domains partially converged, while others were represented mainly by one computational representation. This finding is consistent with recent work showing that computational and researcher-developed qualitative representations may capture related material while differing in interpretation and contextual specificity \citep{misra2026llmqualitative}. The denial and negation theme became unstable after contraction normalization. This instability indicates that preprocessing choices may influence the observed divergence. Human interpretation therefore remains central. Computational models can identify patterns or assign predefined labels, but researchers still need to determine whether outputs are conceptually related and what their relationship means in context. Prior work on topic modeling and qualitative analysis similarly emphasizes linking computational patterns to source material \citep{gillies2022theme}, while recent LLM-based qualitative research continues to place interpretive responsibility with researchers \citep{than2025futurecoding}. PTI supports this process by making alignment decisions and the reasoning behind integrated conclusions traceable. The analysis also distinguished interpretive value from external-anchor association. Baby care and health (T0) and activities and preferences (T2) contributed to the combined external-anchor model. Both were excluded from the joint display, indicating that themes contributing to external-anchor association were not necessarily those selected for integrated qualitative interpretation. Although the combined model achieved a higher observed AUC, this difference should be interpreted cautiously given the small sample and transductive NMF representation. More importantly, this distinction highlights the value of PTI in separating statistical associations from interpretive contributions while making the relationships between computational representations explicit and traceable.

%######################################################################

\subsection{Limitations}\label{subsec:limitations}

The small cohort ($n = 29$, nine depressed cases) limits the strength of the supervised findings, which should be interpreted as exploratory. Findings illustrate the procedure and do not support substantive claims about perinatal mental health. Because NMF was fit on the full corpus before cross-validation, the NMF and combined-model estimates are transductive. NMF solution stability was modest, and the T4 denial and negation theme was not stable after contraction normalization. The PV ontology was developed for clinical dialogue and has not been validated for translated Nepali perinatal narratives. Unmatched themes may therefore reflect a codebook coverage gap, translation effects, labeling error, limited ontology fit, or model instability.

\section{Conclusion}\label{sec:conclusion}

PTI offers a transparent and auditable procedure for integrating two computational representations of one qualitative corpus. One representation uses ontology-guided coding, the other uses data-driven topic modeling. PTI aligns these representations and supports their joint interpretation within mixed methods research. In the perinatal demonstration, the PV representation identified predefined psychosocial constructs, while NMF surfaced additional linguistic patterns. The crosswalk and joint display made inclusion, alignment, and divergence decisions traceable. We treat the two representations as distinct and as capable of being meaningfully aligned and integrated. The demonstration indicates that the two representations may contain partially non-overlapping information that is useful for integration. The balance between shared and representation-specific information may vary across corpora, codebooks, and computational methods. Further applications of PTI to larger and more diverse datasets and substantive domains could examine how such differences influence integrated interpretation and the development of meta-inferences.

\section{Ethical considerations}

This work is a methodological study that uses a secondary analysis of de-identified participant interview transcripts collected by the parent Nepal study. Ethical approval and participant consent for the parent study were obtained by the original investigators, as reported by \citet{vanheerden2024standstrong}. The present analysis used only de-identified text, included quotations without direct identifiers, and involved no new contact with participants.

\section{Data and materials availability}

The de-identified interview data and Patient Voice ontology codebook are available upon request. The analysis code and implementation details are publicly available on GitHub at \url{https://github.com/Elyasirankhah/PTI}.

\begin{acks}
The authors thank the parent study research team and the participants whose interview data made this methodological demonstration possible.
\end{acks}

%\begin{funding}
%[Funding statement to be completed.]
%\end{funding}

\begin{dci}
The author(s) declared no potential conflicts of interest with respect to the research, authorship, and/or publication of this article.
\end{dci}

\bibliographystyle{SageH}
\bibliography{sn-bibliography}

\appendix

% Appendix floats use S-prefixed numbering so figure/table numbers match their "Supplementary" titles.

\setcounter{figure}{0}
\renewcommand{\thefigure}{S\arabic{figure}}

\end{document}